\documentclass[letterpaper, 10 pt, journal, twoside]{IEEEtran}
\usepackage{amsfonts}
\usepackage{algorithmic}
\usepackage{graphicx}
\usepackage{textcomp}
\usepackage{xcolor}
\usepackage{comment}
\usepackage{hyperref}
    \makeatletter
    \let\NAT@parse\undefined
    \makeatother
\usepackage{rotating}
\usepackage{multicol}
\usepackage{multirow}
\usepackage{graphics} 
\usepackage{epsfig} 
\usepackage{times} 
\usepackage{amsmath} 
\usepackage{amssymb}  
\usepackage{cite}
\usepackage{url}
\usepackage{dblfloatfix}    

\usepackage[flushleft]{threeparttable}
\usepackage[pscoord]{eso-pic}
\newcommand{\placetextbox}[3]{
\setbox0=\hbox{#3}
\AddToShipoutPictureFG{ \put(\LenToUnit{#1\paperwidth},\LenToUnit{#2\paperheight}){\vtop{{\null}\makebox[0pt][c]{#3}}}}
}
\placetextbox{.5}{0.035}{~\copyright 2021 IEEE. Personal use of this material is permitted. Permission from IEEE must be obtained for all other uses.}

\begin{document}

\title{
Slip-Based Autonomous ZUPT through Gaussian Process to Improve Planetary Rover Localization
}

\author{Cagri Kilic, Nicholas Ohi, Yu Gu, and Jason N. Gross 
\thanks{Manuscript received: November 23, 2020; Revised January 26, 2021; Accepted March 12, 2021.}
\thanks{This paper was recommended for publication by Editor Pauline Pounds upon evaluation of the Associate Editor and Reviewers' comments.
This work was supported in part by NASA EPSCoR Research Cooperative Agreement WV-80NSSC17M0053, in part by NASA WVSGC Agreement 80NSSC20M0055, and in part by the Benjamin M. Statler Fellowship.} 
\thanks{Authors are with the Department of Mechanical and Aerospace Engineering, West Virginia University, Morgantown, WV 26506 USA 
(E-mail: {\tt\footnotesize cakilic@mix.wvu.edu}, {\tt\footnotesize nohi@mix.wvu.edu}, {\tt\footnotesize yu.gu@mail.wvu.edu}, {\tt\footnotesize jason.gross@mail.wvu.edu})}%
\thanks{Digital Object Identifier (DOI): see top of this page.}}

\markboth{IEEE Robotics and Automation Letters. Preprint Version. Accepted March, 2021}
{Kilic \MakeLowercase{\textit{et al.}}: Slip-Based Autonomous ZUPT through Gaussian Process to Improve Planetary Rover Localization} 

\maketitle

\begin{abstract}

The zero-velocity update (ZUPT) algorithm provides valuable state information to maintain the inertial navigation system (INS) reliability when stationary conditions are satisfied. Employing ZUPT along with leveraging non-holonomic constraints can greatly benefit wheeled mobile robot dead-reckoning localization accuracy. However, determining how often they should be employed requires consideration to balance localization accuracy and traversal rate for planetary rovers. To address this, we investigate when to autonomously initiate stops to improve wheel-inertial odometry (WIO) localization performance with ZUPT. To do this, we propose a 3D dead-reckoning approach that predicts wheel slippage while the rover is in motion and forecasts the appropriate time to stop without changing any rover hardware or major rover operations. We validate with field tests that our approach is viable on different terrain types and achieves a 3D localization accuracy of $\thicksim$97$\textbf{\%}$ over 650~m drives on rough terrain. 

\end{abstract}

\IEEEpeerreviewmaketitle

\section{Introduction}
\label{intro}
\IEEEPARstart{A}{chieving} accurate real-time localization performance is challenging for planetary rovers with limited-performance computers traversing on harsh and unknown terrains that cause wheel slippage. Rover slip is often estimated using visual odometry (VO)~\cite{c54,rankinCuriosity}. Despite its safety and reliability, using VO for long periods comes with some concerns: 1) substantial traversal rate reduction since the rover needs to stop to acquire images~\cite{toupet2019terrain}, and needs to drive slow due to limited computational resources~\cite{li2008characterization}; 2) the low number of detected and tracked features on indistinguishable terrains (e.g., bright areas, sand dunes, shadowed areas) can lead to poor accuracy of motion estimates~\cite{c21} and limit the usage of VO. Specifically, Mars Science Laboratory (MSL) rover reaches a maximum speed of 140 m/h in blind-drive mode (without VO), 45 m/h in hazard avoidance mode (VO update every 10 meters), and only 20 m/h in fully autonomous mode (VO update every half-vehicle length)~\cite{grotzinger2012mars}.  

For current Mars rovers, the slow pace driving can be alleviated by using the blind-driving mode, which makes use of wheel odometry (WO) and inertial measurement unit (IMU) to keep track of the rover's motion if the terrain ahead is considered to be safely traversable by the rover operation team. However, using only blind-driving causes unbounded pose error growth over time and increasing uncertainty in the rover state due to wheel slippage and INS drift. For this reason, the rover localization is corrected with computationally expensive methods after a short period of blind-driving~\cite{c54}. 

Leveraging ``free'' information without affecting any other operations and using observations for multiple purposes are desirable characteristics for planetary missions~\cite{arvidson2003physical}. In planetary missions, stopping is inevitable for the rovers due to hardware constraints, and so far, the autonomous planetary rovers are stopping approximately every 1-10 meters of driving for various reasons~\cite{grotzinger2012mars, toupet2019terrain}. As the rover is mostly stationary due to these frequent stops, ZUPT can be leveraged to maintain INS alignment. The main advantages of ZUPT for the localization task is that it can bound the velocity error, calibrate IMU sensor biases, and limit the rate of INS localization drift~\cite{grovebook}. Using ZUPT in a planetary rover dead-reckoning system can provide a computationally efficient and accurate real-time rover localization capability, even in feature-poor areas, without any major changes to the rover operations. Furthermore, having a more reliable onboard proprioceptive localization approach may help to reduce the frequency of using computationally expensive visual-based corrections. However, knowing how often ZUPT should be employed requires consideration to avoid unnecessarily reducing traverse rate.

In our previous work~\cite{kilic2019improved}, we presented an approach to enhance planetary rover dead-reckoning localization performance by making use of ZUPT with periodic stops. In this study, we propose an autonomous stopping framework by monitoring wheel slippage and predicting the time when the rover needs to stop to keep the localization drift rate to an acceptable level using only an IMU and wheel encoders.
Our contributions are listed as:
\begin{itemize}
    \item We develop a novel method for predicting localization error, using a time-series Gaussian process model for prediction of slip uncertainty as a function of time, such that ZUPTs can be actively initiated with respect to the wheel slippage frequency and magnitude.
    \item We evaluate our approach in a set of field tests and demonstrate that the proposed method is able to improve blind-driving localization on different terrain types (e.g., paved, unpaved, graveled, and rough areas) that yield different stopping times.
    \item We make our software (designed using Robot Operating System~\cite{ROS}), and datasets publicly available in~\cite{gpdata}.
\end{itemize}

The rest of the paper is organized as follows. Section~\ref{related_work} provides a comprehensive overview of related works. In Section~\ref{preliminaries}, we introduce the preliminaries for the problem formulation. In Section~\ref{methodology}, we describe the details of the proposed framework. Section~\ref{experimental_results} explores the concept further and carries out a qualitative analysis of experimental results. Finally, conclusions are presented in Section~\ref{conclusions}.

\section{Related Work}
\label{related_work}

Wheel slippage can occur when the terrain traversed fails~\cite{iagnemma2004mobile} or when there is a kinematic incompatibility between wheels (i.e., different wheel speeds) encountered~\cite{gonzalez2018slippage}.
Because of slippage and imperfect measurement of the wheel radius, WO based localization is inherently subject to drift. 

Knowledge of the terrain geometry is a critical asset for the rovers in unknown environments for safe traversal. For example, MSL uses stereo vision to generate a digital elevation map (DEM) of the surrounding terrain enhanced by leveraging High Resolution Imaging Science Experiment (HiRISE) images~\cite{arvidson2014terrain} similar to Mars Exploration Rovers (MERs)~\cite{rankinCuriosity}. VO is an accurate and reliable source of information for slip estimation; however, it is computationally expensive for planetary rovers. Even with the field-programmable gate array (FPGA) processors~\cite{lentaris2015hw}, the other limitations of VO arise that it suffers from low-feature terrains and it relies on proper lighting conditions~\cite{strader2020perception}. Similarly, insufficiently detected and tracked features may lead to poor accuracy of motion estimate~\cite{c21}. 

Various studies have modeled slip as a function of terrain geometry. Past studies have yielded important insights into the relationship between visual terrain information and the measured slip using training examples by casting the problem into a Mixture of Experts (MoE) framework~\cite{angelova2006learning}. However, this terrain geometry knowledge does not guarantee to localize the rover relative to terrain traversed since the rover slip is measured infrequently, and it causes a substantial reduction of the traversal rate due to computational expenses~\cite{toupet2019terrain}. 

Moreover, the wheel-terrain interactions (terramechanics) are not dictated by the visible topsoil of the terrain~\cite{c21}. To address this, a recent line of research has focused on data-driven cubic regression metrics to predict slip with respect to the slope by using proprioceptive and exteroceptive sensors~\cite{skonieczny2019data}. Although slippage is strongly affected by increasing absolute value of a slope, wheel slippage can also be observed on flat terrains while encountering local obstacles (e.g., small rocks that rover can traverse on) due to kinematic incompatibility~\cite{gonzalez2018slippage}.

Martian soil is extremely challenging for traversability; even throughout a single drive, Mars rovers traverse various terrains~\cite{arvidson2014terrain}. Employing a terramechanics model to estimate slip requires the knowledge of terrain parameters and variables, which are challenging to measure or estimate accurately online. Due to the complexity of terramechanics modeling, considerable research has been devoted to simplified models. For example,~\cite{iagnemma2004mobile} presented a tool for online estimation of terrain parameters based on a simplified terramechanics model for deformable terrains. 

Apart from terramechanics modeling, machine learning algorithms have also been utilized as slip estimation tools. Locally adaptive slip-model learning with respect to slope values is demonstrated in~\cite{cunningham2017locally} using a Gaussian process (GP) regression for visually classified terrain types. Using visual information is one of the common ways to classify a terrain and estimate an equivalent slip value for planetary missions. However, unexpected small variances on the terrain can be deceptive for a vision based slip-learning approach~\cite{brooks2005vibration}. 

The methodology in~\cite{hidalgo2017gaussian} demonstrated an offline wheel slippage learning approach, where the model is learned on training runs and evaluated in a test environment using SLAM in a planetary rover navigating an unstructured environment. On the other hand,~\cite{rogers2012continuous} suggested that the mapping between inputs and resultant behavior depends critically on terrain conditions which vary significantly over time and space (spatio-temporal). Therefore, offline techniques for slip estimation are most likely to suffer from learning changes in wheel-terrain interactions. 

Leveraging ZUPT is a natural fit for wheeled planetary robots because rovers are in stationary conditions in many instances~\cite{biesiadecki2006mars} such as capturing images for obstacle avoidance, re-planning, processing VO, and conducting scientific experiments. When a rover is in stationary conditions, localization performance can be improved by using the pseudo-measurements generated (i.e., ZUPT) as detailed in our previous work~\cite{kilic2019improved}. ZUPT is a well-known concept that was initially popularized to aid inertial pedestrian navigation~\cite{foxlin2005, norrdine2016}. Zero-velocity detection and application on paved  road for automobile applications are shown in~\cite{xiaofang2014,ramo, brossard2019rins}. 

\section{Preliminaries}
\label{preliminaries}

This section introduces several essential framework elements for planetary rover proprioceptive localization from our previous study for the sake of completeness. Detailed descriptions can be found in \cite{kilic2019improved}. 

\subsection{Rover Filter States}
An error state extended Kalman filter (ES-EKF), based on the method detailed in \cite{grovebook}, is implemented to enhance proprioceptive localization and provide uncertainty bounds. The error state vector is formed in a local navigation frame, 
\begin{equation}
\label{errorstate}
\mathbf{x}_{err}^{n}={\biggl(
\delta\mathbf{ \Psi}_{nb}^{n} \ \ 
\mathbf{\delta v}_{eb}^{n} \ \ 
\delta\mathbf{p}_{b} \ \ 
\mathbf{b}_a \ \ 
\mathbf{b}_g
\biggr )}^{\mathbf{T}}
\end{equation}
where, $\delta\mathbf{ \Psi}_{nb}^{n}$ is the attitude error, $\mathbf{\delta v}_{eb}^{n}$ is the velocity error, $\delta\mathbf{p}_{b}$ is the position error, $\mathbf{b}_a$ is the IMU acceleration bias, and $\mathbf{b}_g$ is the IMU gyroscope bias.

It is assumed that the error-state vector is defined by~\eqref{errorstate} and the total state vector is 
\begin{equation}
\mathbf{x}^{n}={\biggl(
\mathbf{\Psi}_{nb}^{n} \ \ 
\mathbf{v}_{eb}^{n} \ \ 
\mathbf{p}_{b} 
\biggr )}^{\mathbf{T}} 
\end{equation}
where each of the nine total states correspond to the first nine error-states.

\subsection{Non-Holonomic Constraints}
A non-holonomic rover is subjected to two motion constraints: 1) zero velocity along the rotation axis of the rover wheels, and 2) zero velocity in the direction perpendicular to the traversed terrain~\cite{diss}. These constraints can be leveraged as a pseudo-measurement update. Assuming that the rear-wheel frame axes are aligned with the body frame, this measurement update can be given as
\begin{equation}
\mathbf{\delta z}_{RC}^{n}=-\begin{pmatrix}
0 & 1 & 0\\ 
0 & 0 & 1
\end{pmatrix} (\mathbf{C}_{n}^{b} \mathbf{v}_{eb}^{n} -\mathbf{\omega}_{ib}^{b} \times \mathbf{L}_{rb}^{b})
\end{equation}
where $\mathbf{C}_{b}^{n}$ is the coordinate transformation matrix from the body frame to the locally level frame, $\mathbf{L}_{br}^{b}$ is body to rear wheel lever arm, and $\mathbf{\omega}_{ib}^{b}$ is angular rate measurement. The approximate measurement matrix can then be found as
\begin{equation}
\mathbf{H}_{RC}^{n}=\begin{pmatrix}
\mathbf{0}_{2,3} & \begin{pmatrix}
0 & 1 & 0 \\ 0 & 0 & 1 \end{pmatrix} \mathbf{C}_{n}^{b} &\mathbf{0}_{2,3}  &\mathbf{0}_{2,3} &\mathbf{0}_{2,3}
\end{pmatrix}
\end{equation}

Note that the lateral velocity constraint is invalid in excessive sideslip conditions. The sideslip angle estimation (see Subsection~\ref{slipdetection}) can be used to verify whether the rover is experiencing an excessive sideslip and this verification can be used to decide the lateral velocity measurement should be omitted or not.   
\subsection{Zero-Velocity Update - (ZUPT) }
During stationary conditions, IMU output is dominated by planetary rotational motion and sensor errors. Therefore, ZUPT can be used to maintain INS accuracy.

ZUPT bounds the velocity error and calibrates IMU sensor biases~\cite{skog}. Hence, the measurement innovation for ZUPT can be expressed as
 \begin{equation}
 \mathbf{\delta z}_{Z,k}^{n -}=[-\mathbf{\hat{v}}_{eb,k}^{n},-\mathbf{\hat{\omega}}_{ib,k}^{b}]^{T} 
 \end{equation}
where $\mathbf{\delta z}_{Z,k}^{n -}$ is measurement innovation matrix, $\mathbf{\hat{v}}_{eb,k}^{n} $ is estimated velocity vector, and $\mathbf{\hat{\omega}}_{ib,k}^{b}$ is estimated gyro bias. The measurement matrix is given as
\begin{equation}
\mathbf{H}_{Z,k}^{n}=\begin{bmatrix}  \mathbf{0}_3 & \mathbf{-I}_3 & \mathbf{0}_3 & \mathbf{0}_3 & \mathbf{0}_3\\
\mathbf{0}_3 & \mathbf{0}_3 & \mathbf{0}_3 & \mathbf{0}_3 & \mathbf{-I}_3
\end{bmatrix}.\end{equation}

\section{Gaussian Process with Time-Series Modeling Overview}
In this study, we employ a GP to model the wheel slippage as time-series data. The primary reason for choosing the GP is to leverage its prediction of uncertainty estimates, which are used for predicting the error-covariance of odometry measurements (see Section~\ref{localization_error}). 

A GP is uniquely defined by its mean function $\mu(x)$ and covariance function $k(x,x')$~\cite{williams2006gaussian}.
\begin{equation}
    f(x) \sim GP(\mu(x),k(x,x'))
\end{equation}
For any collection of input points, $\mathbf{x}=\{x_1,...x_n\}$, with defining a probability distribution $p(f(x_1),...,f(x_n)$, has a joint Gaussian distribution such that
\begin{equation}p\left(f\left(x_{1}\right), \ldots, f\left(x_{n}\right) | x_{1}, \ldots, x_{n}\right)=\mathcal{N}(\boldsymbol{\mu}, \boldsymbol{K})\end{equation}
where the matrix $\boldsymbol{K} \in \mathbb{R}^{n \times n}$ is the kernel matrix whose entries are given by $K_{i j}=k\left(\boldsymbol{x}_{i}, \boldsymbol{x}_{j}\right)$, $i, j=1, \dots, n$, and $\boldsymbol{\mu}$ is the corresponding mean vector. The covariance (kernel) function encodes the similarity between the outputs in GP~\cite{duvenaud2014automatic}. To model the different characteristics of the training dataset, which is collected while the rover is in motion, we combine two kernels as a product to capture the different slip behavior of the rover with respect to the terrain.

Assuming that the slip can be occurred randomly and significantly (e.g., impulsive high slippage) due to unexpected kinematic incompatibility, we adopted the Brownian kernel, $k_B=min(t,t')$. On the other hand, from the mathematical expression of Radial Basis Function (RBF) kernel, $k_{RBF}=\exp (-\left\|t-t^{\prime}\right\|^{2} / 2\ell^{2})$, it can be assumed that if inputs are similar, then the outputs would be similar~\cite{vert2004primer}. In the case that the rover does not encounter significant slippage, we assumed the subsequent measurements should be similar to each other for a short time-interval (the time-interval between two successive slip measurements is 0.1s in our setup) resulting to a repetitive-low slippage. Based on this intuition and a heuristic approach from field test results, we also used RBF kernel in our GP model, resulting in a composite kernel (i.e., multiply kernels together)~\cite{duvenaud2014automatic} such as $k(t,t')=k_B(t,t') k_{RBF}(t,t')$. Note that the assumptions mentioned above are for blind-driving mode, and the mode can be activated when the terrain is considered safe to be driven for planetary rovers. The aim of a regression problem is to learn the mapping from inputs to outputs~\cite{richardson2017gaussian}, given a training set of input and output pairs $\mathbf{(x,y)}={(x_i,y_i)}_{i=1}^{N}$, where $N$ is the number of training examples, predictions can be made at test indices $\mathbf{x_*}$ by computing the conditional distribution and with assuming a zero mean $\epsilon_i \sim \mathcal{N}(0,\sigma_\epsilon^2)$, results in a Gaussian distribution and given by:

\begin{equation}p\left(\mathbf{y}_{*} | \mathbf{x}_{*}, \mathbf{x}, \mathbf{y}\right)=\mathcal{N}\left(\mathbf{y}_{*} | \boldsymbol{\mu}_{*}, \mathbf{\Sigma}_{*}\right)\end{equation}
where 
\begin{align}
  \boldsymbol{\mu}_{*}&=\mathbf{K}_{*}^{T}\mathbf{K}_{*}^{-1} \mathbf{y}, \quad \mathbf{K}_{*}=\mathbf{K}(\mathbf{x},\mathbf{x}_{*}) \\
\mathbf{\Sigma}_{*}&=\mathbf{K}_{**}-\mathbf{K}_{*}^{T}\mathbf{K}_{*}^{-1} \mathbf{K}_{*}, \quad \mathbf{K}_{* *}=\mathbf{K}(\mathbf{x}_{*}, \mathbf{x}_{*}).   
\end{align}

\section{Methodology}
\label{methodology}

\begin{figure}[h!]
\centering
\includegraphics[width=\columnwidth]{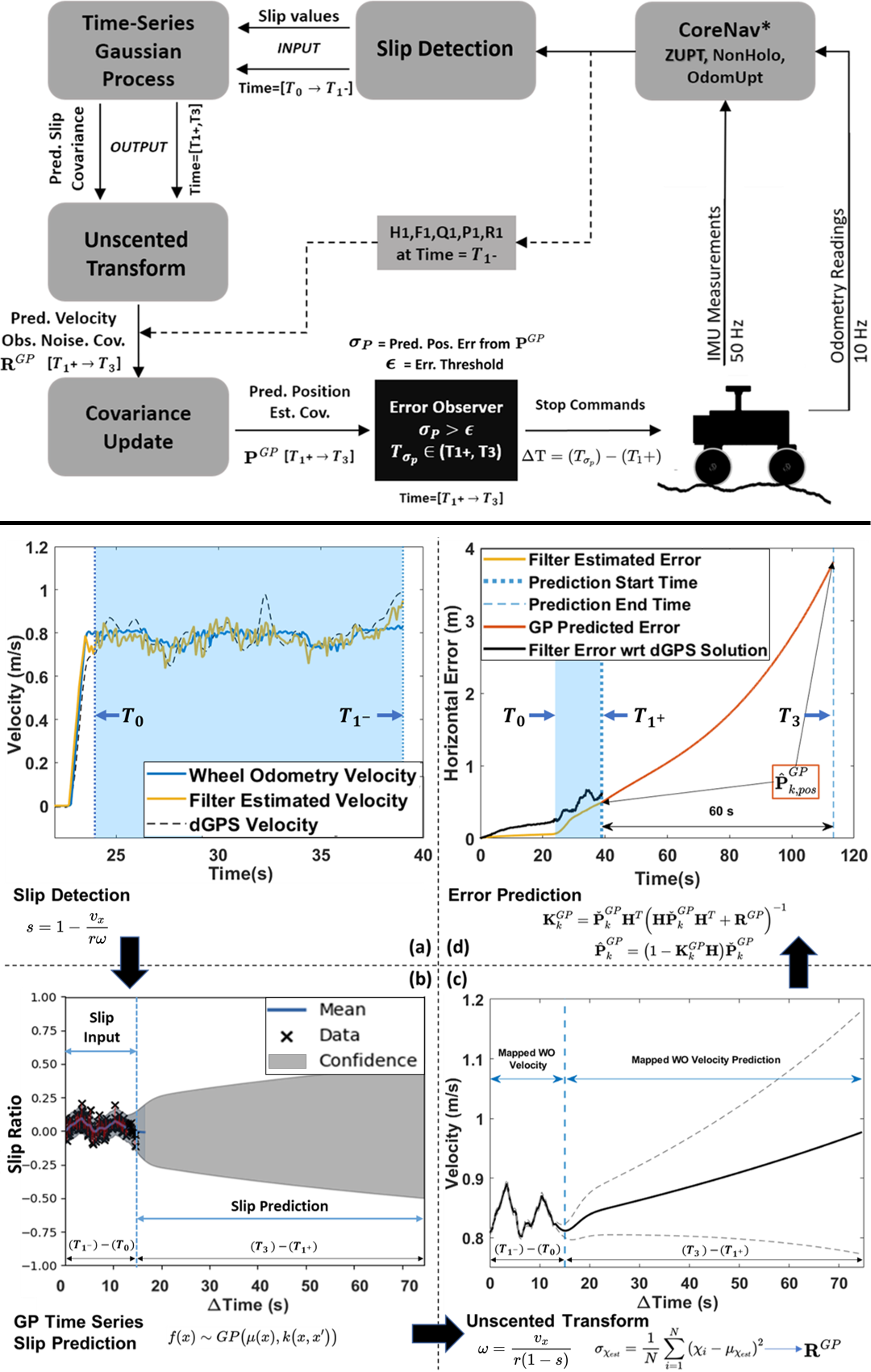}
\caption{The proposed framework is demonstrated on the top figure. Each elements of the framework are shown on the bottom sub-figures. The sub-figure (a) shows the filter estimated and WO estimated velocity to be used for slip detection. The sub-figure (b) shows the slip input and slip prediction. The sub-figure (c) is a depiction of unscented transform that used for mapping the WO velocities for error prediction. Finally, the sub-figure (d) shows how the predicted error is generated. The input slip data is collected within a time window ($[T_0,T_1^-]$) which represented in the blue area in (a). The dotted blue line ($T_1^+$) in sub-figure (d) represents the time when the future error prediction is generated for 60s. The post processed DGPS outputs are assumed as truth and given for comparison purposes. 
}
\label{proofofconcept}
\end{figure}

The proposed wheeled-robot localization framework consists of a series of actions in current-time and future-time, both of which are computed onboard the rover. The current-time portion consists of our previous work~\cite{kilic2019improved}, an INS mechanization aided with WO, pseudo-measurements, and kinematic constraints in an ES-EKF as briefly summarized in Section~\ref{preliminaries}. The future-time part of the framework uses the estimated slip events and prior estimated error state information to predict the robot's localization error. 
A depiction of the proposed framework and its elements is demonstrated in Fig.~\ref{proofofconcept}. 
\subsection{Slip Detection}
\label{slipdetection}

\begin{figure*}[h!]
\centering
\includegraphics[scale=0.4185]{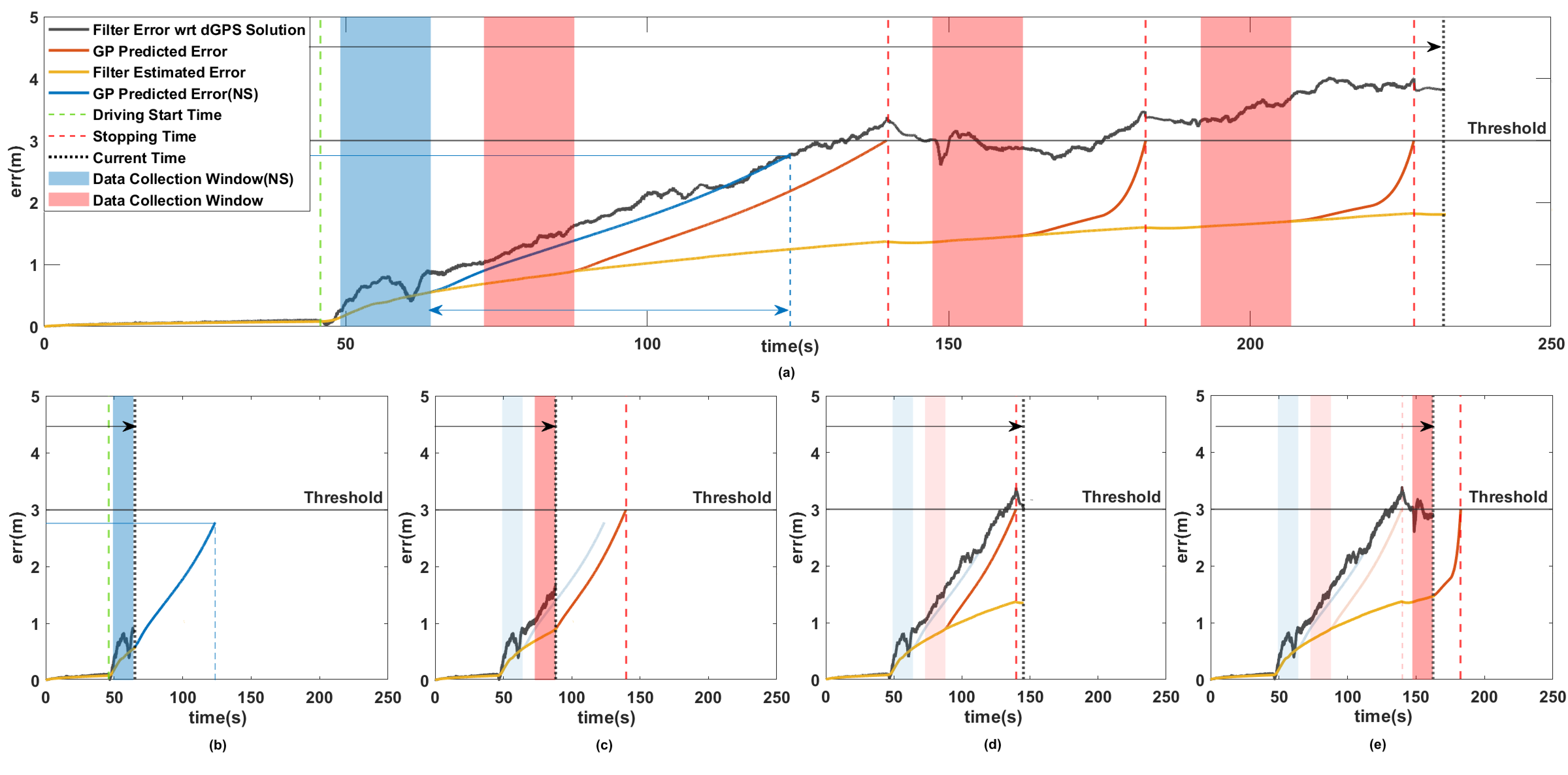}
\caption{ A demonstration of the on-board actions and error prediction process of the proposed algorithm. "Filter Estimated Error" is ES-EKF provided estimation and "Filter Error" is the difference between position truth (post-processed differential GPS solution) and the filter position estimation. The testbed rover's average forward speed is 0.8~m/s. The algorithm only considers the slippage collection from WO and filter estimated velocity for 15s intervals. This interval for learning is set based on engineering judgement to capture the most recent (the last 12m of drive) terrain-wheel information based on the MSL Hazard Avoidance slip check interval (10m)~\cite{grotzinger2012mars}. Threshold is set to 3m and error prediction time limit is set to 60s. The prediction limit is set based on the limitation of blind-driving driving (50m) on MSL operations~\cite{grotzinger2012mars}, and reliability of GP prediction over longer times. (a) Overall error prediction and stopping decision for 230s of the operation. (b) Current time = 64s. After collecting 15s of slippage data, the algorithm predicts the horizontal error for 60s. Since the predicted error does not exceed the predetermined threshold, rover continues driving. (c) Current time = 88s. Same process as (b) but this time the error prediction exceeds the threshold before the 60s prediction time limit. Algorithm sets an internal countdown for stopping at the point when the error prediction exceeds the threshold. (d) Current time = 145s. Rover stops, applies ZUPT, and starts driving again. (e) Current time = 162s. Algorithm collects data for 15s, and predicts the stopping time, repeats the process as (c).  The GP prediction process took less than a second with IntelCore i7-8650U CPU (Intel NUC Board NUC7i7DN) and is negligible to show in the figure.}
\label{exampleScene}
\end{figure*}

The slippage is monitored with the slip ratio calculation for front and rear wheels velocity with respect to the INS velocity. Example estimates of WO based velocity, INS (filter) estimated velocity, and truth (DGPS) velocity are shown in Fig.~\ref{exampleScene} (a). 

The longitudinal slip ratio, $s \in[-1,1]$, is defined as:
\begin{equation}
\label{slip}
s=\left\{\begin{array}{ll}
{1-\frac{v_x}{r \omega}} & { \text { (if }\omega \neq 0, v_x<r \omega,  \text { driving })} \\
{\frac{r \omega}{v_x}-1} & { \text { (if }\omega \neq 0, v_x>r \omega,  \text { braking })}
\end{array}\right.
\end{equation}
where ${v_x}$ is the translational velocity estimated from INS, $r$ is the wheel radius, and $\omega$ is the wheel angular velocity estimated from the WO measurements. The motion estimates from the filter are compared to the computed velocity based on the vehicle kinematics to determine if any slippage has occurred. Detected slippage input is demonstrated in Fig.~\ref{exampleScene} (b). 

Also, sideslip can be expressed using the slip angle, $\mathbf{\beta}$, and can be given as the angle between lateral velocity, $v_y$, and translational velocity $v_x$
\begin{equation}
\mathbf{\beta}=\tan ^{-1}\left(\frac{v_{y}}{v_{x}}\right)
\end{equation}

Although there are several methods to detect slippage as discussed in Section \ref{related_work}, we adopt this proprioceptive slip detection since it is computationally efficient and not required any visual-sensor information to observe the wheel slippage for the proposed method.

\subsection{Wheel Slippage with GP Time-Series Modeling}
In our case, there is one input $(\mathbf{x}=\mathbf{T})$ and one output $(\mathbf{y}=\mathbf{s})$ in the GP. The input $\mathbf{T}=\{t_1,t_2,...t_N\} $ is the time tags of each corresponding slip ratio value, and the output $\{s_1,s_2,...,s_N\}=\mathbf{s} \in [-1,1]$ is the estimated slip ratio value, assuming $N$ training input and output pairs such that $\mathcal{D}=(\mathbf{T},\mathbf{s})$.

The collected training data for wheel slip ratio values, $\mathbf{s}=\{s_1,...,s_N\}$, and corresponding time tags $\mathbf{T}=\{t_1,...,t_N\}$ for a time window are used to learn the model 
\begin{equation}
    s=f(t)+\epsilon, \quad \epsilon \sim \mathcal{N}(0,\sigma^2).
\end{equation}

The time window for learning is kept short to capture the most current (the last 12~m of drive) terrain-wheel information based on the MSL Hazard Avoidance slip check interval ($\sim$10~m) \cite{grotzinger2012mars}.
In that time window, the rover is in free driving (i.e., rover does not perform any stops). The learned model is then processed in the GP forecast model to make predictions at future test indices $\mathbf{t}_{*}=\left\{t_{{*}_{i}}\right\}_{i=1}^{N^{+}}$ for future unknown wheel slip ratio observations $\mathbf{s}_{*}=\left\{s_{{*}_{i}}\right\}_{i=1}^{N^{+}}$ where ${N^{+}}$ is the number of test indices which in our case it corresponds to a future time tag. For a detailed demonstration of slip input and slip prediction by using the slip ratio definition, see Fig.~\ref{proofofconcept}(b). A python GP library \cite{gpy2014} is used in our rover's ROS framework to optimize the hyperparameters (e.g., the length parameter $l$ in the RBF kernel), and to predict the slip values while the rover is in motion. 

\subsection{Wheel Odometry Velocity Prediction}
\label{unscented_transform}
To predict the simulated odometry velocity error boundaries, a statistical sigma point transformation inspired by unscented transformation~\cite{julier2000new} where the slip ratio definition in~(\ref{slip}) is used to generate this transformation function:

\begin{equation}
{\sigma}_{{\chi_{e s t}}_{(t)}} = \frac{1}{N} \sum_{i=1}^{N}\left(\chi_{{i}_{(t)}}-\mu_{{\chi_{e s t}}_{(t)}}\right)^{2}, \quad t \in [ T_1^+, T_3 ]
\end{equation}

where ${T_1^+}$ is the time when the prediction is being generated, $T_3$ is the time when the generated prediction ends (i.e., $T_{3}=T_1+60$~s, see Fig.~\ref{proofofconcept}(c)),  $N$ is the number of the sigma points, $\chi_i$ is velocity term mapped from slip measurement, defined as $\chi_1={\mu_{vel}}/{(1-\mu_s)}$, $\chi_{2}={\mu_{vel}}/{(1-\mu_s - \sigma_s)}$, and $\chi_{3}={\mu_{vel}}/{(1-\mu_s + \sigma_s)}$ where $\mu_s$ and $\sigma_s$ are mean and variance of $s$, respectively, and $\chi_{est}$ is the mean of $\chi_i$ values for $N=3$. 

In constituting the observation noise covariance matrix in the localization forecasting phase,~$\mathbf{R}^{GP}$, we assumed that the constant WO velocity related $\mathbf{R}$ values on the filter could be interchangeable with varying $\sigma_{\chi_{est}}$ values between ${T_1^+}$ and ${T_3}$ come from predicted observation covariance.

\begin{equation}
\mathbf{R}^{GP} = \begin{bmatrix}
 (\mathbf{\sigma}_{\chi_{est}}^2\mathbf{I}_{3x3})& \mathbf{0} \\ 
\mathbf{0}&1 \\  \end{bmatrix}_{4x4}
\end{equation}

These mapped velocity values and their prediction with this statistical sigma point transformation method are depicted in Fig.~\ref{proofofconcept}(c). 

\subsection{Forecasting Localization Error}
\label{localization_error}
When the forecasted GP data arrives, the algorithm uses the latest filter error covariance estimate, $\mathbf{P_{T_1^-}}$, to initialize the error covariance prediction. 
\begin{equation}
\hat{\mathbf{P}}_{0}^{G P} = \mathbf{P_{T_1^-}}    
\end{equation}
The most recent state transition matrix, $\mathbf{F_{T_1^-}}$, process noise covariance, $\mathbf{Q_{T_1^-}}$, and WO observation matrix, $\mathbf{H}_{T_1^-}$ are being kept fixed during the forecasting error covariance process (see the left side of the Fig.~\ref{proofofconcept}).
\begin{equation}
\check{\mathbf{P}}_{k}^{G P}=\mathbf{F_{T_1^-}} \hat{\mathbf{P}}_{k-1}^{G P} \mathbf{F_{T_1^-}}^{T}+\mathbf{Q_{T_1^-}}
\end{equation}

Then, the algorithm simulates an INS error covariance propagation. In our setup, simulated odometry update is assumed to take place in every 5th IMU time step (IMU data rate is 50 Hz, WO data rate is 10 Hz). When this simulated odometry update is available, transformation function predicts the simulated odometry velocity error boundaries. Finally, the simulated Kalman gain is calculated and simulated estimate covariance is updated. 

\begin{equation}
\mathbf{K}_{k}^{G P}=\check{\mathbf{P}}_{k}^{G P} \mathbf{H_{T_1^-}}^{T}\left(\mathbf{H_{T_1^-}} \check{\mathbf{P}}_{k}^{G P} \mathbf{H_{T_1^-}}^{T}+\mathbf{R}^{G P}\right)^{-1}
\end{equation}

\begin{equation}
\hat{\mathbf{P}}_{k}^{G P}=\left(1-\mathbf{K}_{k}^{G P} \mathbf{H_{T_1^-}}\right) \check{\mathbf{P}}_{k}^{G P}
\end{equation}

For each updated covariance prediction, the algorithm calculates the position error covariances as a function of time. An example calculation is illustrated in Fig.~\ref{proofofconcept}(d). When the horizontal error gets more prominent than a predetermined threshold, the algorithm takes the corresponding time for that event, calculates the remaining time to stop with respect to the current time, and alerts the rover to stop. If there is no need for stopping (e.g., the positioning error prediction is below the threshold within the prediction time limit), the rover keeps driving. Otherwise, the rover stops traversing, applies ZUPT, then keeps driving. A detailed example scenario is given Fig.~\ref{exampleScene}. The details to model state transition matrix $\mathbf{F}$, process noise covariance $\mathbf{Q}$, observation matrix $\mathbf{H}$ and the observation noise covariance $\mathbf{R}$ can be found in \cite{kilic2019improved} and \cite{grovebook}.

     \begin{figure}[h!]
      \centering
      \includegraphics[scale=0.26]{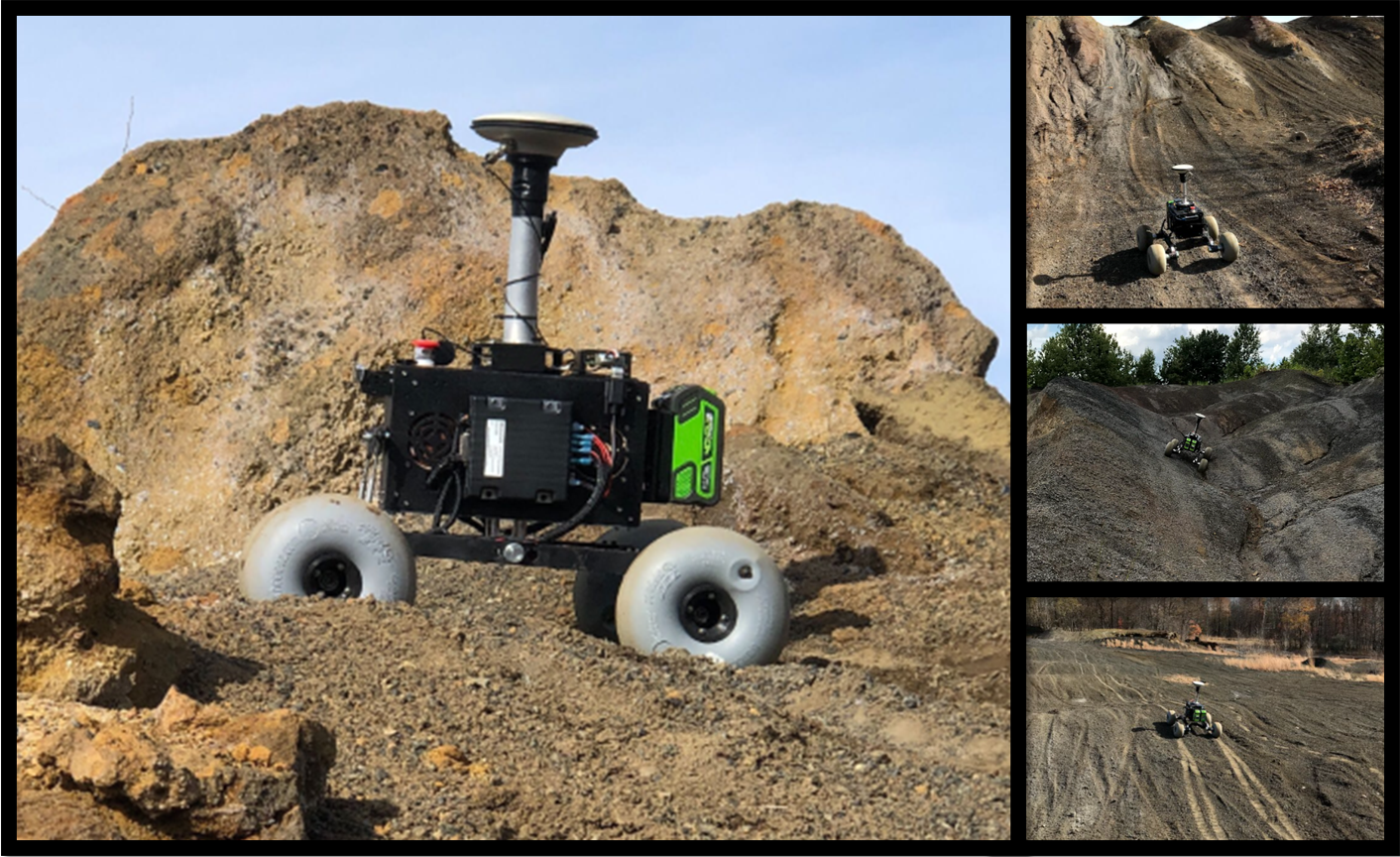}
      \caption{Pathfinder test platform during field tests in Point Marion, PA. }
      \label{pathfinder}
      \end{figure} 
\section{Experimental Results}
\label{experimental_results}
\subsection{Setup}
Pathfinder, a custom-built testbed rover, is employed for the experimental evaluation of the proposed method (see Fig.~\ref{pathfinder}). The platform is a lightweight, 4-wheeled, skid-steered robot. Rover uses a rocker system with a differential bar connected to the front wheels. In general, planetary rovers use wheels with grousers, which increase traction and traversability performance (e.g., MSL, MERs, ExoMars). However, Pathfinder is utilized with slick wheels to test our localization algorithm against significant slippage. Slick wheels lead to encounter more slippage with larger frequency and occurrence which aid to detect slippage but degrade the localization performance significantly.

The IMU used on the rover is an ADIS-16495 with 50~Hz data rate~\cite{adis} and the quadrature encoders are used for WO readings with 10~Hz data rate. Integer-ambiguity-fixed carrier-phase differential GPS (DGPS) is used to determine a truth reference solution. Just as in~\cite{kilic2019improved}, dual-frequency Novatel GPS receivers and L1/L2 Pinwheel antennas~\cite{novatel1} are mounted to the rover and a stationary base station. During the experiments, 10~Hz carrier-phase and GPS pseudorange measurements were logged on both receivers.  Rover state is initialized with a loosely-coupled GPS-IMU sensor fusion algorithm, such as driving straight with a short distance ($\sim$10~m) for estimating initial heading and being stationary for a period of time ($\sim$30~s) to initialize position before testing. After initialization, GPS measurements are collected only externally for generating the truth through post-processing. The open-source software library, RTKLIB 2.4.2~\cite{rtklib}, is used to post-process the DGPS solutions with a cm-to-dm expected level accuracy~\cite{gps}. Rover is teleoperated and commanded for 0.8~m/s forward speed in field tests.

\subsection{Evaluation}
A series of tests were performed on several terrains, including paved, unpaved, gravel, and rough terrains. Paved terrains are relatively flat roads with minimal slippage observation. 
      \begin{figure}[h!]
      \centering
      \includegraphics[scale=0.26]{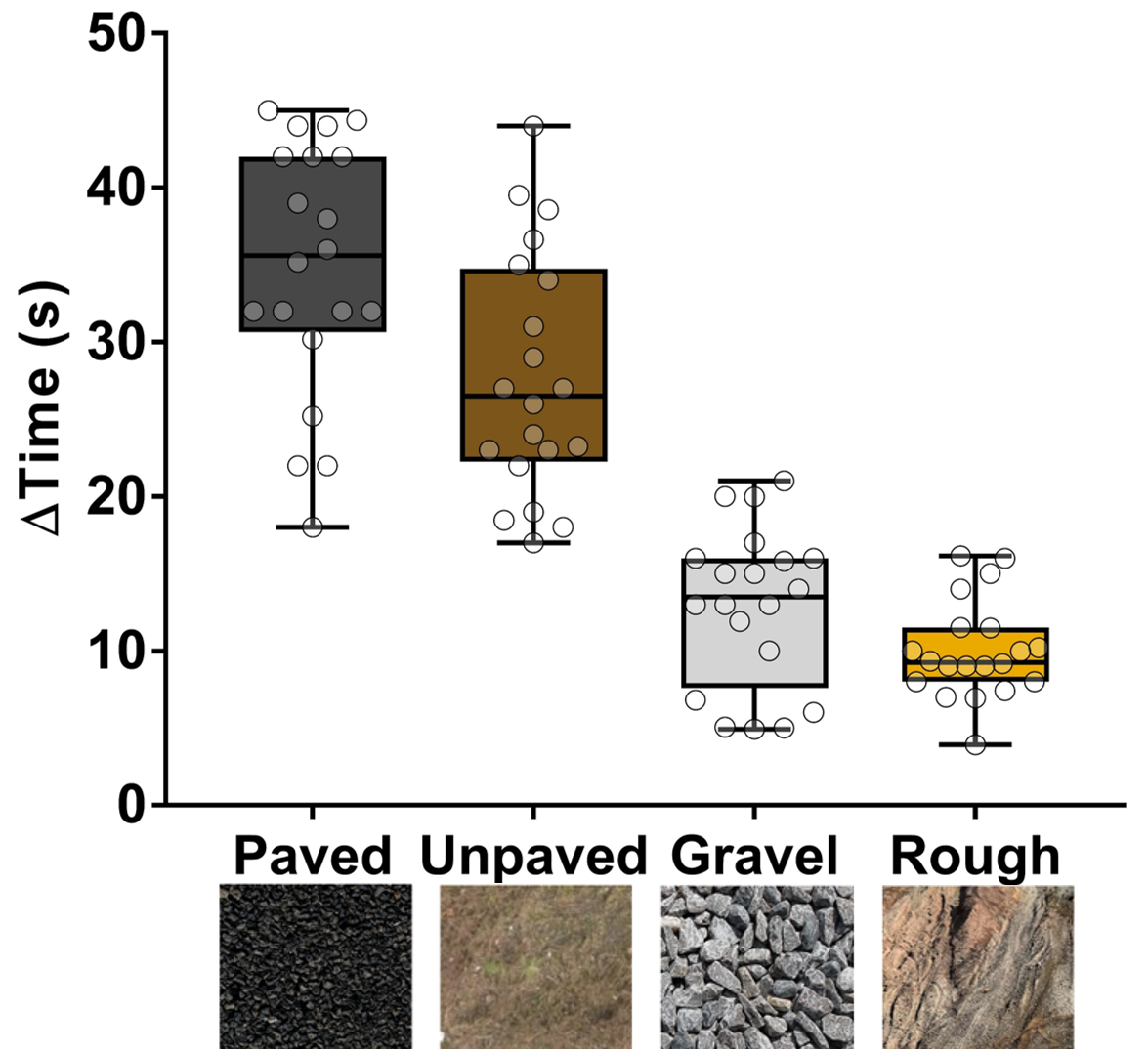}
    \caption{Comparison of stop time interval for terrain types. $\Delta$Time axis in box plot shows the remaining duration to stop after 15s of data collection. The middle line in the boxes show the median value of 20 tests for each terrain type. GraphPad Prism software v.7 is used for one-way analysis of variance Tukey's multiple comparison statistical analysis test.  Paved: P, Unpaved: U, Gravel: G Rough: R. Non-significant difference: G/R (\textit{p}=0.5028). Significant difference: P/U (\textit{p}=0.0049), R/P, R/U, G/P, and G/U (\textit{p}$<$0.0001).}
      \label{deltatime}
      \end{figure}Unpaved terrains are also rigid roads with small scattered rocks that rover can easily traverse. Gravel terrain consists of small broken rock materials. Due to the shape of these materials, there is less traction on the wheels on the gravel road. This loose surface creates slippage primarily due to wheel kinematic incompatibilities. 
This letter particularly focuses on the rough terrain results because of its similarities with the Martian terrain (see Fig.~\ref{pathfinder}). This terrain is a burnt coal ash pile located at Point Marion, PA, with complex geometric (e.g., sloped, pitted, fractured, and sandy areas) and chemical terrain properties~\cite{ramme2004we} similar to the abundant chemical compounds found in Martian regolith~\cite{peters2008mojave}. 

A stopping time comparison analysis against four terrain types is shown in Fig.~\ref{deltatime}. In this analysis, the rover is driven on different terrains and the corresponding stop time intervals are stored. Paved and unpaved roads are rigid, and the terrain underneath the wheels is not moving, and the robot wheels do not encounter significant slippage, resulting in better WO. However, the rover encounters significant slippage on gravel (kinematic incompatibility) and rough terrain (sinkage, slope, and kinematic incompatibility). The important result of this analysis is that the average stopping time intervals are shorter on gravel and rough terrain than on benign roads. Correspondingly, the algorithm enforces the rover stops more often on more slippery terrains  (minimum stop frequency is~15~s).

To further evaluate the method, the localization accuracy of the proposed estimation is compared against the DGPS solution. As detailed in Table~\ref{tab:results}, we achieved approximately 1\% of 3D localization error (ENU) in short (152~m) and medium (339~m) range distances on rough terrain with keeping the stopping error threshold as 2~m. Also, in long~(650~m) range distances, the threshold is varied as 2~m, 3~m, and 5~m to observe the localization accuracy performance against stopping time prediction. In these field test results, the algorithm reasonably predicts the stopping time to keep the localization drift approximately 3$\%$ for the 5~m threshold and less than 2$\%$ for the 3~m threshold. We also monitored that the rover often does not need to stop for the 5~m threshold due to not exceeding the threshold in the prediction time limit.

\begin{table} [h!]
\centering
\footnotesize
\begin{threeparttable}
\caption{Accuracy of the Proposed Approach on Rough Terrain}
\label{tab:results}
\centering
\begin{tabular}{@{}lcccccc@{}}
\hline
 Ash Pile & \multicolumn{4}{c}{Test Specifics\tnote{*}}   &\multicolumn{2}{c}{Error ($\%$) }\\
& \scriptsize{$\Sigma_{D}$(m)}& \scriptsize{$\epsilon$(m)}& \scriptsize{Stop Count}& \scriptsize{$\Sigma_{T}$(s)}& \scriptsize{ENU}& \scriptsize{Median}\\ 
\hline\hline
Test1          &671	&5	&8	&879	&3.07	&1.73\\
Test2          &663	&3	&19	&924	&1.78	&1.04\\
Test3          &652	&3	&20	&915	&1.14	&1.05\\
Test4        &339	&2	&9	&469	&0.91	&0.58\\
Test5         &152	&2	&5	&215	&0.94	&0.82\\

\hline
 & \multicolumn{3}{c}{Horizontal Error (m)}   &\multicolumn{3}{c}{RMS Error (m)}\\
 & \scriptsize{Median}& \scriptsize{STD}& \scriptsize{Max.}& \scriptsize{East}& \scriptsize{North}& \scriptsize{Up}\\ 
\hline\hline
Test1          &11.60	&12.01	&34.48	&17.84	&7.26	&7.25\\
Test2          &6.89	&5.68	&18.78	&9.49	&3.08	&6.24\\
Test3          &6.84	&2.88	&11.10	&5.67	&4.44	&1.75\\
Test4        &1.96	&1.27	&5.86	&1.74	&1.83	&1.78\\
Test5         &1.24   &0.80   &2.72   &1.34   &0.47   &0.13\\
\hline
\end{tabular}
\begin{tablenotes}
\item[*] $\Sigma_{D}$: Traversed Distance, $\epsilon$: Error Threshold, $\Sigma_{T}$: Traversal Time.
\end{tablenotes}
\end{threeparttable}
\end{table}
    \begin{figure}[h!]
      \centering
      \includegraphics[scale=0.169]{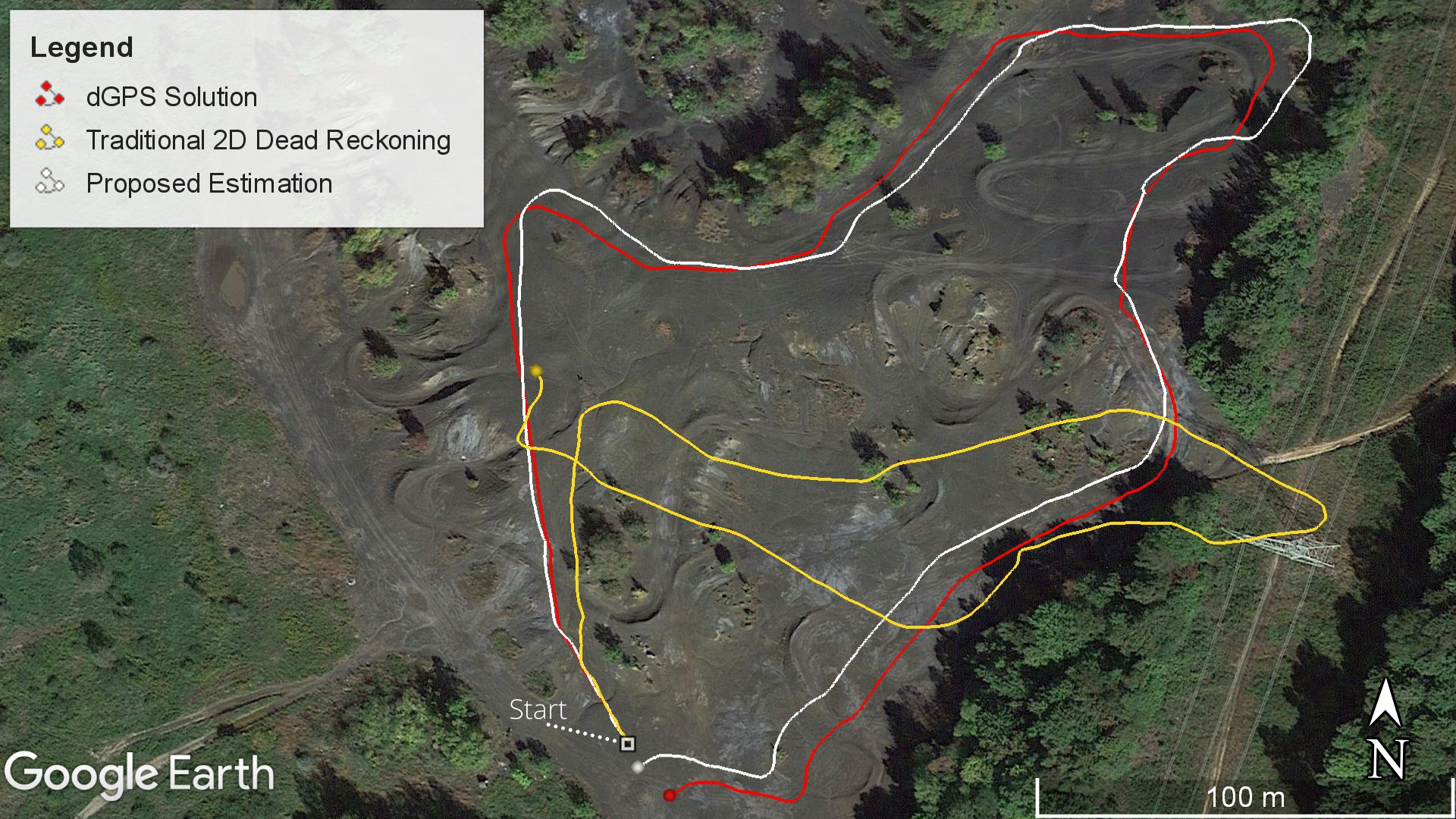}
      \caption{Ash-Pile test result for 652m driving with 3m error threshold.}
      \label{ashpile}
      \end{figure}
Ground-track depiction of an example scenario from ash-pile field testing is given in Fig.~\ref{ashpile}. 
The results show that traditional 2D dead-reckoning (WIO) is reliable only for short distances due to slippage, whereas the proposed estimation (3D~WIO+ZUPT) can be used for longer distances if the terrain is safe to drive blindly.
The localization design goal for MER was to maintain a position estimate that drifted less than 10$\%$ during a 100~m drive~\cite{c54}. Without using ZUPT and kinematic constraints in blind-driving, the drift can quickly elevate and exceed that design limit, as shown in Fig.~\ref{ashpile}.

Moreover, a comparison analysis between autonomous (proposed) and periodic~\cite{kilic2019improved} stopping methods is provided in Table~\ref{tab:comparison}. Using autonomous stopping leads to an average stop rate $(S_R$) decrease over 65\% compared to periodic stopping while keeping the localization accuracy more than 98\%. Consequently, when using ZUPT, autonomous stopping increases the traversal rate by stopping less, and keeps the localization accuracy to an acceptable level.
\begin{table} [t]
\centering
\footnotesize
\begin{threeparttable}
\caption{Periodic versus Autonomous ZUPT Comparison}
\label{tab:comparison}
\centering
\begin{tabular}{@{}lccccc@{}}
\hline
 Periodic & \scriptsize{$\Sigma_{D}$(m)}& \scriptsize{$\Sigma_{T}$(s)}& \scriptsize{Error(\%)}& \scriptsize{$S_{R}$(\%)}&  \scriptsize{$\Sigma$Stop}\\ 
\hline\hline
Rough\_A         &151	&504	&0.85	&25.02	 &42\\
Unpaved\_A          &87	&133	&1.53	&18.08		 &8\\
Unpaved\_B         &128	&181	&1.02	&11.60		 &7\\
\hline
Autonomous & \scriptsize{$\Sigma_{D}$(m)}& \scriptsize{$\Sigma_{T}$(s)}& \scriptsize{Error(\%)}& \scriptsize{$S_R$(\%)}&   \scriptsize{$\Sigma$Stop}\\ 
\hline\hline
Rough\_B          &152	&215	&0.94	&7.32		&5\\
Unpaved\_C          &183	&244	&1.17	&6.15		&5\\
Unpaved\_D         &161	&210	&1.56	&4.28		&3\\
\hline
\end{tabular}
\begin{tablenotes}
\item[*] $\Sigma_{D}$, $\Sigma_{T}$: Same as Table~\ref{tab:results}, $S_R$: Stop rate, $\Sigma$Stop: Stop count
\end{tablenotes}
\end{threeparttable}
\end{table}
     \begin{figure}[h!]
      \centering
      \includegraphics[width=\columnwidth]{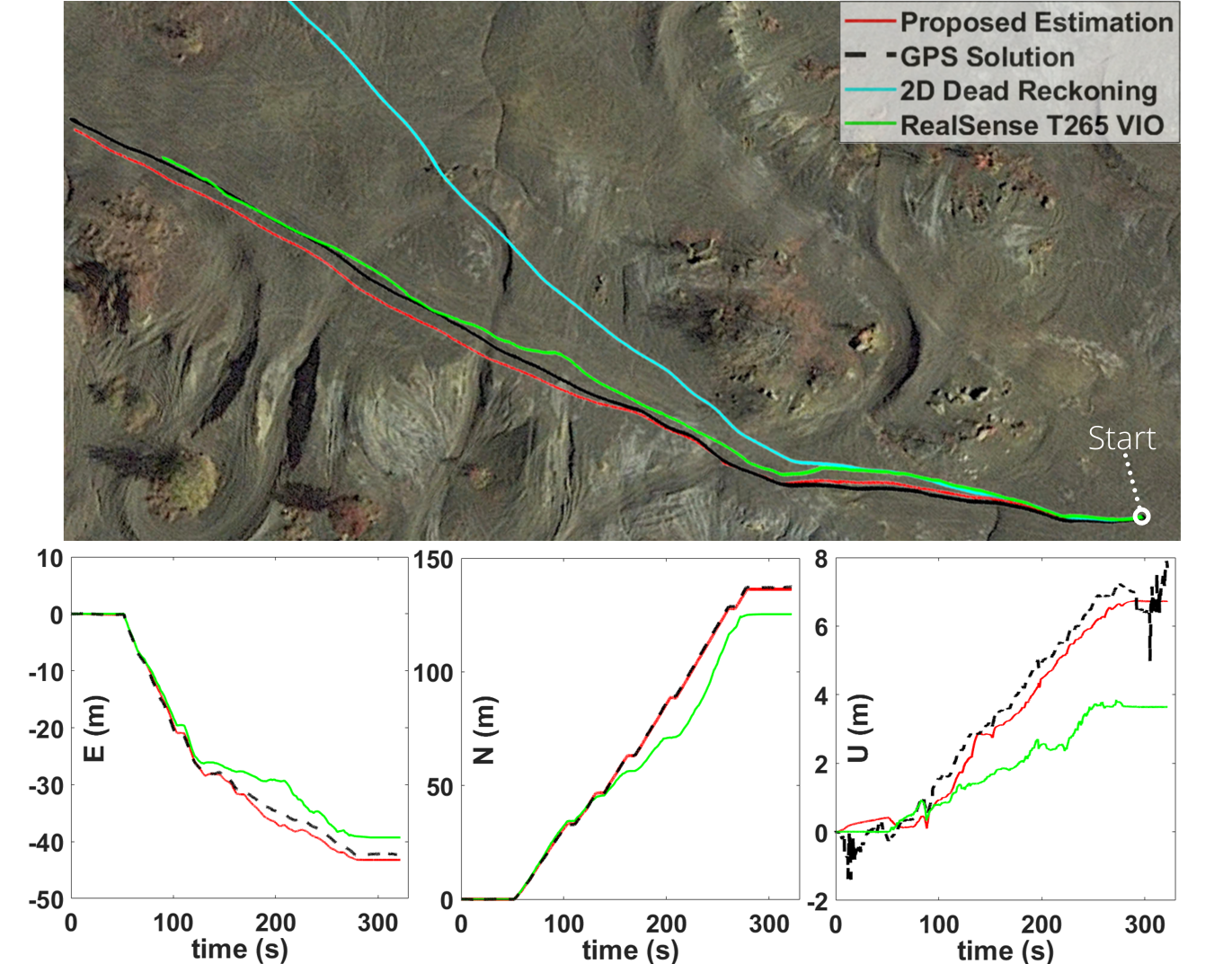}
      \caption{Depiction of a comparison for localization accuracy of the proposed approach (3D~WIO+ZUPT) in a low-feature rough terrain against RealSense T265~\cite{intel} VIO, and 2D dead-reckoning (WIO). Traversed distance 150m. RMSE VIO  = East: 2.70m, North: 10.41m, Up: 2.12m. RMSE Proposed = East: 1.03m, North: 0.49m, Up: 0.65m. A detailed analysis with several other examples is available in the VIO Analysis folder at \url{https://github.com/wvu-navLab/CN-GP}.  }
      \label{placeholder}
      \end{figure}
      
A comparison between our localization approach against a commercially off-the-shelf RealSense T265 tracking system~\cite{intel}  visual-inertial odometry (VIO) solution is provided in Fig.~\ref{placeholder}. In this field test, rover traversed for 150~m on a low-feature terrain. The tracking system is able to provide reliable solution in feature rich areas whereas it suffers in the areas with a lack of detectable and trackable features. This is a common issue of visual-based localization approaches because these approaches require reasonable distinct visual features in view to operate accurately~\cite{campos2020orb,strader2020perception,rankinCuriosity}.

\section{Conclusion and Future Work}
\label{conclusions}

We presented a slip-based localization error prediction framework, which effectively balances the traversal-rate and localization accuracy for wheeled planetary rovers. Instead of periodic stopping, ZUPTs can be autonomously initiated with respect to the wheel slippage frequency and magnitude using a time-series GP model for prediction of slip uncertainty as a function of time. Planetary robot slip related localization drift can be alleviated with ZUPTs and can provide reliable localization performance for longer distances. The main value of the proposed approach is that it can be easily integrated into planetary rover operations (and many other wheeled robots) to improve onboard localization performance with no hardware changes and minimal operational changes. Since planetary rovers are already stopping frequently for using VO or other operational reasons, using ZUPT along with the blind-drive is a natural fit. 

Future work will focus on 1) using deformable planetary spring tires with a traction control mechanism to help alleviating the limitation of the method when stopping on a steep slope and sliding down, 2) improving the method with adaptive and robust filtering techniques. 

Collected dataset for experimental validation is available in~\cite{gpdata} for the community to use. Developed software and supplementary analyses for this paper are available at:\\
\url{https://github.com/wvu-navLab/CN-GP}.

\addtolength{\textheight}{-2cm}   

\bibliographystyle{IEEEtran}
\bibliography{references}

\begin{thebibliography}{10}
\providecommand{\url}[1]{#1}
\csname url@samestyle\endcsname
\providecommand{\newblock}{\relax}
\providecommand{\bibinfo}[2]{#2}
\providecommand{\BIBentrySTDinterwordspacing}{\spaceskip=0pt\relax}
\providecommand{\BIBentryALTinterwordstretchfactor}{4}
\providecommand{\BIBentryALTinterwordspacing}{\spaceskip=\fontdimen2\font plus
\BIBentryALTinterwordstretchfactor\fontdimen3\font minus
  \fontdimen4\font\relax}
\providecommand{\BIBforeignlanguage}[2]{{%
\expandafter\ifx\csname l@#1\endcsname\relax
\typeout{** WARNING: IEEEtran.bst: No hyphenation pattern has been}%
\typeout{** loaded for the language `#1'. Using the pattern for}%
\typeout{** the default language instead.}%
\else
\language=\csname l@#1\endcsname
\fi
#2}}
\providecommand{\BIBdecl}{\relax}
\BIBdecl

\bibitem{c54}
M.~Maimone, Y.~Cheng, and L.~Matthies, ``Two years of visual odometry on the
  mars exploration rovers,'' \emph{Journal of Field Robotics}, vol.~24, no.~3,
  pp. 169--186, 2007.

\bibitem{rankinCuriosity}
A.~{Rankin}, M.~{Maimone}, J.~{Biesiadecki}, N.~{Patel}, D.~{Levine}, and
  O.~{Toupet}, ``Driving curiosity: Mars rover mobility trends during the first
  seven years,'' \emph{2020 IEEE Aerospace Conference}, pp. 1--19, 2020.

\bibitem{toupet2019terrain}
O.~Toupet, J.~Biesiadecki, A.~Rankin, A.~Steffy, G.~Meirion-Griffith,
  D.~Levine, M.~Schadegg, and M.~Maimone, ``Terrain-adaptive wheel speed
  control on the curiosity mars rover: Algorithm and flight results,''
  \emph{Journal of Field Robotics}, 2019.

\bibitem{li2008characterization}
R.~Li, B.~Wu, K.~Di, A.~Angelova, R.~E. Arvidson, I.-C. Lee, M.~Maimone, L.~H.
  Matthies, L.~Richer, R.~Sullivan \emph{et~al.}, ``Characterization of
  traverse slippage experienced by spirit rover on husband hill at gusev
  crater,'' \emph{Journal of Geophysical Research: Planets}, vol. 113, no. E12,
  2008.

\bibitem{c21}
R.~Gonzalez and K.~Iagnemma, ``Slippage estimation and compensation for
  planetary exploration rovers. state of the art and future challenges,''
  \emph{Journal of Field Robotics}, vol.~35, no.~4, pp. 564--577, 2018.

\bibitem{grotzinger2012mars}
J.~P. Grotzinger, J.~Crisp, A.~R. Vasavada, R.~C. Anderson, C.~J. Baker,
  R.~Barry, D.~F. Blake, P.~Conrad, K.~S. Edgett, B.~Ferdowski \emph{et~al.},
  ``Mars science laboratory mission and science investigation,'' \emph{Space
  science reviews}, vol. 170, no. 1-4, pp. 5--56, 2012.

\bibitem{arvidson2003physical}
R.~Arvidson, R.~Anderson, A.~Haldemann, G.~Landis, R.~Li, R.~Lindemann,
  J.~Matijevic, R.~Morris, L.~Richter, S.~Squyres \emph{et~al.}, ``Physical
  properties and localization investigations associated with the 2003 mars
  exploration rovers,'' \emph{Journal of Geophysical Research: Planets}, vol.
  108, no. E12, 2003.

\bibitem{grovebook}
P.~D. Groves, \emph{Principles of GNSS, inertial, and multisensor integrated
  navigation systems}.\hskip 1em plus 0.5em minus 0.4em\relax Artech house,
  2013.

\bibitem{kilic2019improved}
C.~Kilic, J.~N. Gross, N.~Ohi, R.~Watson, J.~Strader, T.~Swiger, S.~Harper, and
  Y.~Gu, ``Improved planetary rover inertial navigation and wheel odometry
  performance through periodic use of zero-type constraints,'' in \emph{2019
  IEEE/RSJ International Conference on Intelligent Robots and Systems (IROS)},
  Nov 2019, pp. 552--559.

\bibitem{ROS}
M.~Quigley, K.~Conley, B.~Gerkey, J.~Faust, T.~Foote, J.~Leibs, R.~Wheeler, and
  A.~Y. Ng, ``Ros: an open-source robot operating system,'' in \emph{ICRA
  workshop on open source software}, vol.~3, no. 3.2.\hskip 1em plus 0.5em
  minus 0.4em\relax Kobe, Japan, 2009, p.~5.

\bibitem{gpdata}
\BIBentryALTinterwordspacing
C.~Kilic and J.~N. Gross, ``Pathfinder gps, imu, and wheel odometry data on
  various terrains,'' 2020. [Online]. Available:
  \url{https://dx.doi.org/10.21227/vz7z-jc84}
\BIBentrySTDinterwordspacing

\bibitem{iagnemma2004mobile}
K.~Iagnemma and S.~Dubowsky, \emph{Mobile robots in rough terrain: Estimation,
  motion planning, and control with application to planetary rovers}.\hskip 1em
  plus 0.5em minus 0.4em\relax Springer Science \& Business Media, 2004,
  vol.~12.

\bibitem{gonzalez2018slippage}
R.~Gonzalez, D.~Apostolopoulos, and K.~Iagnemma, ``Slippage and immobilization
  detection for planetary exploration rovers via machine learning and
  proprioceptive sensing,'' \emph{Journal of Field Robotics}, vol.~35, no.~2,
  pp. 231--247, 2018.

\bibitem{arvidson2014terrain}
R.~Arvidson, P.~Bellutta, F.~Calef, A.~Fraeman, J.~B. Garvin, O.~Gasnault,
  J.~A. Grant, J.~Grotzinger, V.~Hamilton, M.~Heverly \emph{et~al.}, ``Terrain
  physical properties derived from orbital data and the first 360 sols of mars
  science laboratory curiosity rover observations in gale crater,''
  \emph{Journal of Geophysical Research: Planets}, vol. 119, no.~6, pp.
  1322--1344, 2014.

\bibitem{lentaris2015hw}
G.~Lentaris, I.~Stamoulias, D.~Soudris, and M.~Lourakis, ``Hw/sw codesign and
  fpga acceleration of visual odometry algorithms for rover navigation on
  mars,'' \emph{IEEE Transactions on Circuits and Systems for Video
  Technology}, vol.~26, no.~8, pp. 1563--1577, 2015.

\bibitem{strader2020perception}
J.~Strader, K.~Otsu, and A.-a. Agha-mohammadi, ``Perception-aware autonomous
  mast motion planning for planetary exploration rovers,'' \emph{Journal of
  Field Robotics}, vol.~37, no.~5, pp. 812--829, 2020.

\bibitem{angelova2006learning}
A.~Angelova, L.~Matthies, D.~Helmick, G.~Sibley, and P.~Perona, ``Learning to
  predict slip for ground robots,'' in \emph{Proceedings 2006 IEEE
  International Conference on Robotics and Automation, 2006. ICRA 2006.}\hskip
  1em plus 0.5em minus 0.4em\relax IEEE, 2006, pp. 3324--3331.

\bibitem{skonieczny2019data}
K.~Skonieczny, D.~K. Shukla, M.~Faragalli, M.~Cole, and K.~D. Iagnemma,
  ``Data-driven mobility risk prediction for planetary rovers,'' \emph{Journal
  of Field Robotics}, vol.~36, no.~2, pp. 475--491, 2019.

\bibitem{cunningham2017locally}
C.~Cunningham, M.~Ono, I.~Nesnas, J.~Yen, and W.~L. Whittaker,
  ``Locally-adaptive slip prediction for planetary rovers using gaussian
  processes,'' in \emph{2017 IEEE International Conference on Robotics and
  Automation (ICRA)}.\hskip 1em plus 0.5em minus 0.4em\relax IEEE, 2017, pp.
  5487--5494.

\bibitem{brooks2005vibration}
C.~A. Brooks and K.~Iagnemma, ``Vibration-based terrain classification for
  planetary exploration rovers,'' \emph{IEEE Transactions on Robotics},
  vol.~21, no.~6, pp. 1185--1191, 2005.

\bibitem{hidalgo2017gaussian}
J.~Hidalgo-Carri{\'o}, D.~Hennes, J.~Schwendner, and F.~Kirchner, ``Gaussian
  process estimation of odometry errors for localization and mapping,'' in
  \emph{2017 IEEE International Conference on Robotics and Automation
  (ICRA)}.\hskip 1em plus 0.5em minus 0.4em\relax IEEE, 2017, pp. 5696--5701.

\bibitem{rogers2012continuous}
F.~Rogers-Marcovitz, N.~Seegmiller, and A.~Kelly, ``Continuous vehicle slip
  model identification on changing terrains,'' in \emph{RSS 2012 Workshop on
  Long-term Operation of Autonomous Robotic Systems in Changing
  Environments}.\hskip 1em plus 0.5em minus 0.4em\relax Citeseer, 2012.

\bibitem{biesiadecki2006mars}
J.~J. Biesiadecki, E.~T. Baumgartner, R.~G. Bonitz, B.~Cooper, F.~R. Hartman,
  P.~C. Leger, M.~W. Maimone, S.~A. Maxwell, A.~Trebi-Ollennu, E.~W. Tunstel
  \emph{et~al.}, ``Mars exploration rover surface operations: Driving
  opportunity at meridiani planum,'' \emph{IEEE robotics \& automation
  magazine}, vol.~13, no.~2, pp. 63--71, 2006.

\bibitem{foxlin2005}
E.~Foxlin, ``Pedestrian tracking with shoe-mounted inertial sensors,''
  \emph{IEEE Computer graphics and applications}, no.~6, pp. 38--46, 2005.

\bibitem{norrdine2016}
A.~Norrdine, Z.~Kasmi, and J.~Blankenbach, ``Step detection for zupt-aided
  inertial pedestrian navigation system using foot-mounted permanent magnet,''
  \emph{IEEE Sensors Journal}, vol.~16, no.~17, pp. 6766--6773, 2016.

\bibitem{xiaofang2014}
L.~Xiaofang, M.~Yuliang, X.~Ling, C.~Jiabin, and S.~Chunlei, ``Applications of
  zero-velocity detector and kalman filter in zero velocity update for inertial
  navigation system,'' in \emph{Proceedings of 2014 IEEE Chinese Guidance,
  Navigation and Control Conference}.\hskip 1em plus 0.5em minus 0.4em\relax
  IEEE, 2014, pp. 1760--1763.

\bibitem{ramo}
A.~Ramanandan, A.~Chen, and J.~A. Farrell, ``Inertial navigation aiding by
  stationary updates,'' \emph{IEEE Transactions on Intelligent Transportation
  Systems}, vol.~13, no.~1, pp. 235--248, 2012.

\bibitem{brossard2019rins}
M.~Brossard, A.~Barrau, and S.~Bonnabel, ``Rins-w: Robust inertial navigation
  system on wheels,'' \emph{arXiv preprint arXiv:1903.02210}, 2019.

\bibitem{diss}
G.~Dissanayake, S.~Sukkarieh, E.~Nebot, and H.~Durrant-Whyte, ``The aiding of a
  low-cost strapdown inertial measurement unit using vehicle model constraints
  for land vehicle applications,'' \emph{IEEE transactions on robotics and
  automation}, vol.~17, no.~5, pp. 731--747, 2001.

\bibitem{skog}
I.~Skog, P.~Handel, J.-O. Nilsson, and J.~Rantakokko, ``Zero-velocity
  detection—an algorithm evaluation,'' \emph{IEEE transactions on biomedical
  engineering}, vol.~57, no.~11, pp. 2657--2666, 2010.

\bibitem{williams2006gaussian}
C.~K. Williams and C.~E. Rasmussen, \emph{Gaussian processes for machine
  learning}.\hskip 1em plus 0.5em minus 0.4em\relax MIT press Cambridge, MA,
  2006, vol.~2, no.~3.

\bibitem{duvenaud2014automatic}
D.~Duvenaud, ``Automatic model construction with gaussian processes,'' Ph.D.
  dissertation, University of Cambridge, 2014.

\bibitem{vert2004primer}
J.-P. Vert, K.~Tsuda, and B.~Sch{\"o}lkopf, ``A primer on kernel methods,''
  \emph{Kernel methods in computational biology}, vol.~47, pp. 35--70, 2004.

\bibitem{richardson2017gaussian}
R.~R. Richardson, M.~A. Osborne, and D.~A. Howey, ``Gaussian process regression
  for forecasting battery state of health,'' \emph{Journal of Power Sources},
  vol. 357, pp. 209--219, 2017.

\bibitem{gpy2014}
{GPy}, ``{GPy}: A gaussian process framework in python,''
  \url{http://github.com/SheffieldML/GPy}, 2012.

\bibitem{julier2000new}
S.~Julier, J.~Uhlmann, and H.~F. Durrant-Whyte, ``A new method for the
  nonlinear transformation of means and covariances in filters and
  estimators,'' \emph{IEEE Transactions on automatic control}, vol.~45, no.~3,
  pp. 477--482, 2000.

\bibitem{adis}
\emph{{ADIS16495} Data Sheet}, Analog Devices, 2017, rev. A.

\bibitem{novatel1}
\emph{{OEM615 Receivers} Data Sheet}, Novatel, 2015, ver.8.

\bibitem{rtklib}
T.~Takasu, ``Rtklib: Open source program package for rtk-gps,''
  \emph{Proceedings of the FOSS4G}, 2009.

\bibitem{gps}
P.~Misra and P.~Enge, ``Global positioning system: signals, measurements and
  performance second edition,'' \emph{Massachusetts: Ganga-Jamuna Press}, 2006.

\bibitem{ramme2004we}
B.~W. Ramme and M.~P. Tharaniyil, \emph{We energies coal combustion products
  utilization handbook}.\hskip 1em plus 0.5em minus 0.4em\relax We Energies,
  2004.

\bibitem{peters2008mojave}
G.~H. Peters, W.~Abbey, G.~H. Bearman, G.~S. Mungas, J.~A. Smith, R.~C.
  Anderson, S.~Douglas, and L.~W. Beegle, ``Mojave mars
  simulant—characterization of a new geologic mars analog,'' \emph{Icarus},
  vol. 197, no.~2, pp. 470--479, 2008.

\bibitem{intel}
``Intel realsense tracking camera,'' \emph{Product Documentation}, 2020.

\bibitem{campos2020orb}
C.~Campos, R.~Elvira, J.~J.~G. Rodr{\'\i}guez, J.~M. Montiel, and J.~D.
  Tard{\'o}s, ``Orb-slam3: An accurate open-source library for visual,
  visual-inertial and multi-map slam,'' \emph{arXiv preprint arXiv:2007.11898},
  2020.

\end{thebibliography}

\end{document}